\documentclass[lettersize,journal]{IEEEtran}
\usepackage{amsmath,amsfonts}
\usepackage{algorithmic}
\usepackage{algorithm}
\usepackage{array}
\usepackage{textcomp}
\usepackage{stfloats}
\usepackage{url}
\usepackage{verbatim}
\usepackage{graphicx}
\usepackage{cite}
\usepackage{subcaption}

\usepackage{mathrsfs}
\usepackage{multirow}
\usepackage{makecell}
\usepackage{pifont} 
\usepackage{xcolor}
\usepackage[pagebackref,breaklinks,colorlinks]{hyperref}

\usepackage{xfrac} 
\usepackage[normalem]{ulem}
\usepackage{times}
\usepackage{soul}
\usepackage[font=small]{caption}
\usepackage{amsthm}
\usepackage[switch]{lineno}
\usepackage[normalem]{ulem}

\usepackage{placeins}
\usepackage{booktabs}
\usepackage{extarrows}
\usepackage{pgfplots} 
\pgfplotsset{compat=1.18} 
\usepackage[table]{xcolor}
\usepackage{nicematrix}

\definecolor{lightred}{RGB}{255, 230, 230} 
\definecolor{lightgreen}{RGB}{235, 255, 235}
\definecolor{verylightgray}{gray}{0.95}

\begin{document}

\title{Counterfactual Reasoning for Fine-Grained Evidence Disentanglement in VideoQA}

\author{IEEE Publication Technology,~\IEEEmembership{Staff,~IEEE,}
\thanks{This paper was produced by the IEEE Publication Technology Group. They are in Piscataway, NJ.}
\thanks{Manuscript received April 19, 2021; revised August 16, 2021.}}

\author{
        Zhou Du,
        Hamid Krim,
        Xiao Wu,
        Zhaoquan Yuan,
        Liangwei Li,
        and~Keisuke Fujii
\thanks{
Corresponding author: Keisuke Fujii.}
\thanks{
{Zhou Du and Keisuke Fujii are with the Graduate School of Informatics, Nagoya University, Nagoya 464-8601, Japan (e-mail: du.zhou@g.sp.m.is.nagoya-u.ac.jp; fujii@i.nagoya-u.ac.jp).

Hamid Krim is with the Department of Electrical and Computer Engineering, North Carolina State University, Raleigh, NC 27695, USA (e-mail: ahk@ncsu.edu).

Xiao Wu and Zhaoquan Yuan are with the School of Computing and Artificial Intelligence, Southwest Jiaotong University, Chengdu 611730, China (email: wuxiaohk@gmail.com; zqyuan@swjtu.edu.cn)

Liangwei Li is with School of OptoElectonic Science and Engineering, University of Electronic Science and Technology of China, Chengdu 610031, China (e-mail: vnlee@std.uestc.edu.cn)
}
}
}

\markboth{IEEE TRANSACTIONS ON MULTIMEDIA}%
{Shell \MakeLowercase{\textit{et al.}}: A Sample Article Using IEEEtran.cls for IEEE Journals}

\maketitle

\begin{abstract}
Recent advances in video multimodal models have significantly improved VideoQA performance. However, these systems often rely on spurious statistical correlations rather than answer-relevant causal evidence, resulting in unfaithful and brittle reasoning, especially in complex real-world scenarios. Existing methods either rely on cross-modality correlations, costly curated training resources, or insufficient causal assumptions and constraints, and typically operate at the time-interval level. As a result, they fail to explicitly disentangle causal visual cues from confounders and provide limited fine-grained evidence localization. 
To address this issue, we propose a \textbf{C}ounterfactual \textbf{R}easoning framework for fine-grained \textbf{E}vidence \textbf{Di}sentanglemen\textbf{T} (CREDiT). CREDiT formulates the VideoQA process using a structural causal model and learns cross-modality representations that are explicitly decomposed into causal and non-causal components under independence and minimality constraints. To facilitate faithful disentanglement, we introduce feature-level causal interventions and construct counterfactual inputs that approximate causal effects while suppressing non-causal correlations. 
Extensive experiments on NExT-GQA, SportsQA, and SPORTU-video demonstrate that CREDiT consistently improves answer accuracy and reasoning reliability across both generic and complex sports scenarios, leading to more trustworthy VideoQA systems.
\end{abstract}


%
\IEEEpeerreviewmaketitle

\section{Introduction}
%
%
%
%

\IEEEPARstart{V}{ideo} question answering (VideoQA) necessitates models that can predict accurate answers based on the input videos and corresponding questions. Recent advances in Video Multimodal Large Language Models (MLLMs)~\cite{lin2023videollava, wang2024tarsier} have significantly improved question-answering (QA) performance by leveraging large-scale pretraining and cross-modality alignment.

\begin{figure}[t]
  \centering
  \captionsetup[subfigure]{skip=2pt} 
  \begin{subfigure}{1\columnwidth}
    \centering
    \includegraphics[width=0.87\linewidth]{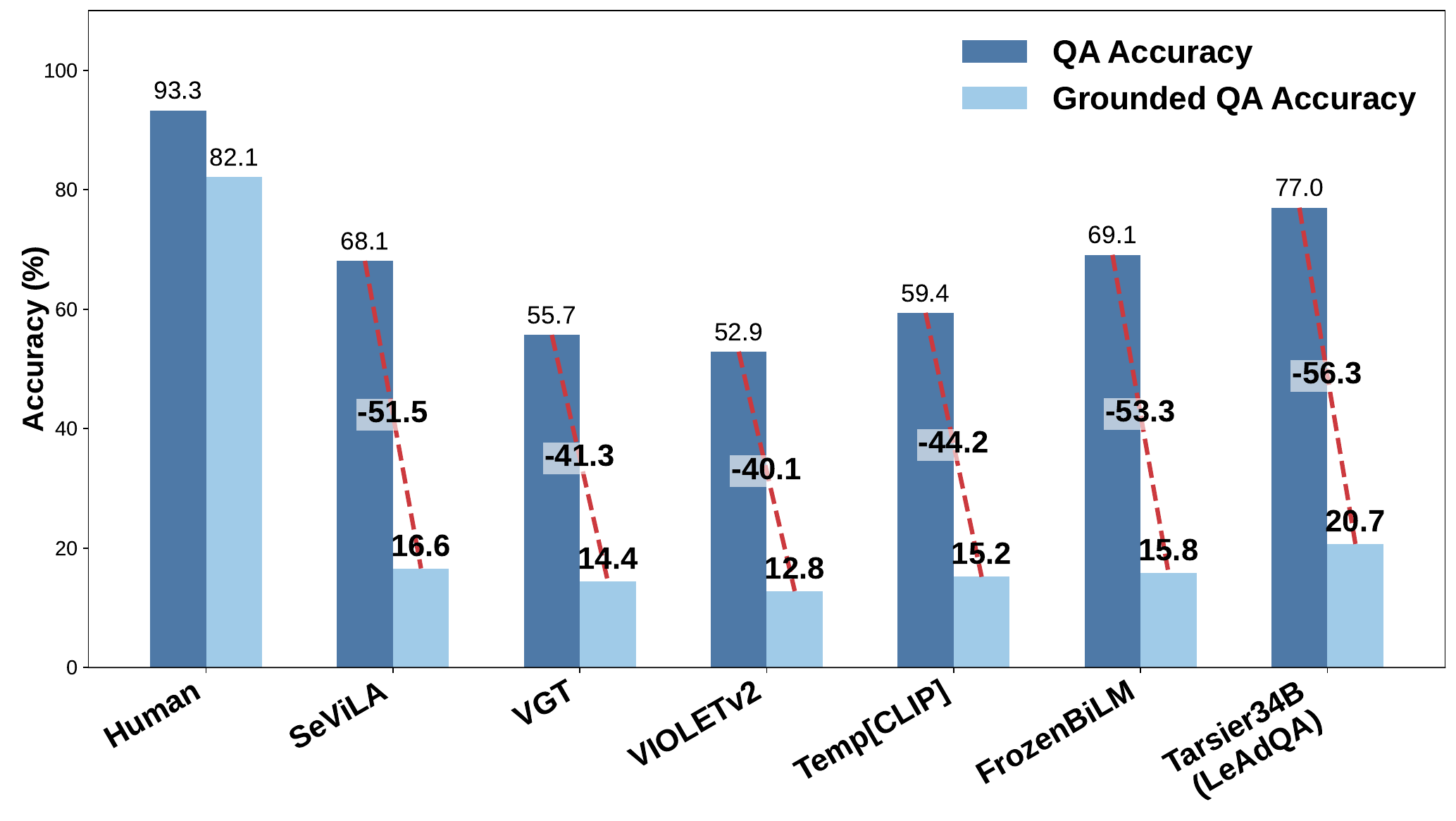}
    \vspace{-4pt}
    \caption{}
    \label{fig:trust_gap}
  \end{subfigure}
  
  \vspace{5pt} 
  \begin{subfigure}{0.87\columnwidth}
    \centering
    \includegraphics[width=\linewidth]{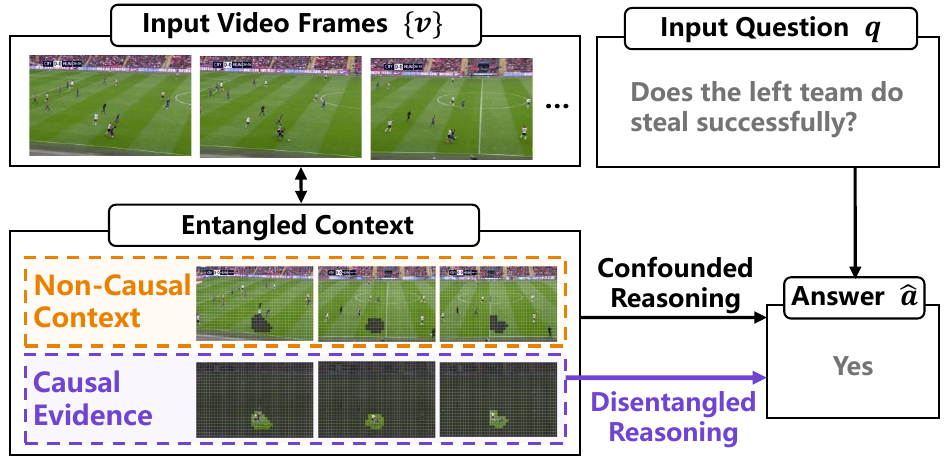}
    \vspace{-8pt}
    \caption{}
    \label{fig:motivation}
  \end{subfigure}
  \vspace{-2pt} 
  \caption{(a) Performance gap between QA accuracy and grounded QA accuracy on NExT-GQA, indicating that existing models are often right without relying on correct visual evidence. The results are reported in~\cite{xiao2024nextgqa,dong2025leadqa}. (b) Illustration of faithful VideoQA reasoning, where answers are derived from disentangled causal visual content rather than possible spurious correlations.}
  \label{fig:combined_figure}
  \vspace{-17pt}
\end{figure}

Although existing models can achieve high answer accuracy, their reasoning processes remain vulnerable. As illustrated in Fig.~\ref{fig:trust_gap}, these models exhibit a significant performance collapse when required to ground their answers in accurate visual evidence.
A compelling explanation for this trustworthiness gap is that most existing methods rely on the Empirical Risk Minimization (ERM) paradigm, which directly minimizes the loss between predictions and ground truth. Under this paradigm, models are prone to capturing all statistical dependencies~\cite{arjovsky2019invariant} rather than learning to identify the key video content, which often causes their attention to become entangled with causally irrelevant context~\cite{chen2025crosscra,xiao2024nextgqa}, leading to the confounded reasoning illustrated in Fig.~\ref{fig:motivation}. 
Such a defect leads to unfaithful inference and potential performance degradation in real-world scenarios with diverse spatiotemporal dynamics~\cite{li2024sportsqadatset,kim2024videoicl}.
In contrast, to achieve faithful reasoning, models should first identify causal visual evidence for the QA process before generating answers.

Although recent studies have attempted to address these challenges, several limitations remain. 
(1) Many approaches identify visual evidence in a correlation-driven manner, relying primarily on statistical or semantic relevance, such as vision-language retrieval~\cite{zhang2024simplellovi} and matching~\cite{yang2022tubedetr,qian2024videostreaming,Wang_2025core,gupta2025toga}. However, such correlations do not necessarily reflect causal evidence and may lead to fragile reasoning under dataset biases or confounding contexts. 
(2) Several recent methods achieve strong grounding performance by leveraging curated manual annotations~\cite{chen2022groundingvqa} and large-scale pretraining~\cite{yu2023SeViLAself,NEURIPS2025_VideoChatR1.5}. However, this reliance may limit their robustness in out-of-domain scenarios~\cite{Chaudhari_2024}, while such costly training resources are rarely available for most VideoQA datasets. 
(3) A few studies attempt to incorporate causal reasoning into VideoQA through linear interpolation-based constraints~\cite{li2022equivariant} or heuristic confounder proxies~\cite{chen2025crosscra}, yet they still lack explicit interventions to sufficiently disentangle non-causal dependencies. Moreover, existing approaches rarely explore beyond temporal interval grounding, leaving fine-grained visual evidence disentanglement underexplored.
These limitations motivate the development of a more fine-grained and trustworthy VideoQA framework that can generalize to realistic open-ended scenarios without requiring additional training costs.

Inspired by the human reasoning process and the robustness of causal relations among variables, we propose a novel \textbf{C}ounterfactual \textbf{R}easoning framework for fine-grained \textbf{E}vidence \textbf{Di}sentanglemen\textbf{T} (CREDiT) in VideoQA. 
We first formulate the correlations in the VideoQA reasoning process via a Structural Causal Model (SCM). Based on this formulation, we design a disentanglement module. This module performs spatio-temporal encoding and textual feature integration to capture complex dynamics and fuse cross-modality semantics, and further disentangles the causal and non-causal components under two explicit constraints: (i) independence, which reduces statistical dependence, and (ii) minimality, which prevents the leakage of non-causal signals. To promote faithful disentanglement, we introduce feature-level interventions on variables in the SCM. Specifically, we construct counterfactual inputs by selectively intervening on causal and non-causal variables, enabling the model to (i) capture answer-relevant causal cues, and (ii) isolate the spurious effect of non-causal factors. Consequently, the proposed method enforces causal evidence capture during the QA process, leading to more robust and interpretable reasoning. Moreover, unlike prior work, CREDiT does not rely on additional annotations, making it applicable to broad VideoQA scenarios.
The key contributions of our work can be summarized as follows:
\begin{itemize}
    \item We present an in-depth formulation of the VideoQA reasoning from a causality-aware perspective to uncover the roots of spurious correlations, alongside a cross-modality causal disentanglement method that decomposes representations into causal and non-causal components under explicit independence and minimality constraints.
    \item We propose a novel intervention-based learning framework that conducts counterfactual reasoning through feature-level interventions on underlying causal and non-causal variables, 
    thereby enabling finer-grained disentanglement of answer-relevant visual evidence.
    \item Extensive experiments demonstrate the effectiveness of CREDiT. CREDiT achieves Acc@QA of $70.4\%$, $60.4\%$, and $71.9\%$ on NExT-GQA, Sports-QA, and SPORTU-video, respectively, demonstrating strong improvement across both generic and complex sports scenarios. Moreover, the improvements in grounded QA accuracy on NExT-GQA, together with further analyses, indicate that CREDiT enables more accurate causal evidence capture for reasoning.
\end{itemize}

\section{Related Work}

\subsection{Video Question Answering}
Video Question Answering (VideoQA) conventionally aims to answer natural language questions based on video content, requiring joint understanding of visual semantics, temporal dynamics, and language reasoning~\cite{jang2017tgif}. Classical VideoQA methods~\cite{gao2018motionappearance,fan2019HMEheterogeneous,jiang2020reasoninghga} primarily relied on recurrent architectures and attention mechanisms to model temporal information and align visual-textual representations. With the emergence of more challenging benchmarks~\cite{xiao2021nextqadataset,li2024mvbenchVideoChat2}, the research focus has gradually shifted toward more complex abilities, including temporal and causal reasoning.
More recently, Transformer~\cite{vaswani2017transformer} architectures and large-scale pretraining have become the dominant paradigm. Existing VideoQA methods adopt diverse Transformer designs~\cite{fu2021violet,xiao2022videovgt}, and leverage vision-language pretraining to improve the multimodal representation~\cite{fu2021violet, yu2023SeViLAself}. Benefiting from the stronger human priors and instruction tuning, recent advances in MLLMs, such as Video-LLaMA~\cite{lin2023videollava}, VideoChatGPT~\cite{maaz2023videochatgpt}, and Qwen2.5-VL~\cite{2025Qwen2.5vl}, substantially improve VideoQA performance on challenging benchmarks.

Despite remarkable progress in QA accuracy, recent studies~\cite{xiao2024nextgqa} reveal that existing VideoQA models often fail to ground their predictions in true visual evidence. Instead, they tend to rely on statistical shortcuts, raising concerns about the faithfulness and trustworthiness of current VideoQA systems. 

\subsection{Video Temporal Grounding and Grounded VideoQA}

Video Temporal Grounding (VTG) aims to localize temporal moments in videos that semantically correspond to language queries~\cite{gao2017tallactivity}. Early VTG methods~\cite{xu2019QSPNmultilevel} follow a two-stage paradigm that generates temporal proposals and performs cross-modality matching, while later Transformer-based models~\cite{moon2023queryQDDETR} formulate VTG as a temporal boundary localization problem and demonstrate superior performance through cross-modality interaction.

Unlike VTG, grounded VideoQA requires not only evidence localization but also complex reasoning ability~\cite{xiao2024nextgqa}. Early work such as TVQA+~\cite{lei2020tvqa+} provides paired annotations between words and bounding boxes, which is closer to word-level grounding than reasoning evidence capture.
Recent approaches, including CoRe~\cite{Wang_2025core}, TOGA~\cite{gupta2025toga}, and structured reasoning methods~\cite{fei2024videovot,Liu_2025commonsense}, perform visual grounding via semantic similarity. Language-based methods~\cite{zhang2024simplellovi} convert videos into texts for retrieval and reasoning. However, these cross-modality correlations often fail to reflect true causal evidence, leading to fragility under spurious statistical biases.
Another line of work improves performance through additional supervision or pretraining. Some models~\cite{wang2024groundedvideollm,dong2025leadqa} rely on ground-truth temporal annotations for grounding learning. Pretraining-based models~\cite{yu2023SeViLAself,fei2024videovot} depend on large-scale video-text data. Notably, by exploiting the MLLM backbone, VideoChat~\cite{NEURIPS2025_VideoChatR1.5} achieves SOTA performance through both curated supervision and targeted multi-task pretraining. However, such costly resources are often unavailable in real-world VideoQA, limiting model generalization~\cite{Chaudhari_2024}.

Several studies have also attempted to model the invariant causal relationships in VideoQA. EIGV~\cite{li2022equivariant} identifies causal clips via contrastive learning but is limited by interpolation-based operations and constraints that cannot explicitly remove spurious dependencies. CRA~\cite{chen2025crosscra} uses cluster centers as heuristic proxies for confounders, yet they still cannot adequately represent real-world confounding factors. Furthermore, existing methods predominantly focus on temporal evidence capture while lacking more granular grounding capabilities.

Motivated by these developments and limitations, our study targets answer-relevant visual evidence disentanglement and aims to develop a more fine-grained trustworthy VideoQA approach that avoids reliance on curated annotations or costly pretraining and can generalize to realistic scenarios.


\section{Preliminaries}

\subsection{Problem Statement}
VideoQA can be formalized as a visual question answering paradigm with sequential inputs. Given an input video $v$ and a question $q$, the model $f_\theta(\cdot)$ predicts an answer $\hat{a}$ that best approximates the ground-truth answer $a$. The objective is to select the answer with the highest probability:
\begin{equation}\label{equation:VideoQA示意公式}
\hat{a} = \arg\max_{a} P_{\theta}(a \mid v, q).
\end{equation}

\subsection{A Causal View of VideoQA}

Current research often approaches causality in data through diverse observational and inferential methods to construct a systematic causal framework, enabling interpretation and prediction of phenomena beyond statistical associations~\cite{pearl2009causal}.
Motivated by this perspective, we introduce a task-driven SCM abstraction for the VideoQA reasoning process that characterizes the underlying causal relationships, identifying the sources of confounded reasoning and guiding the design of corresponding solutions.
As illustrated in Fig.~\ref{fig:causality_videoqa}, the video $V$ and question $Q$ jointly give rise to the cross-modality representation $M$, i.e., $V,Q \rightarrow M$. Subsequently, $M$ serves as the evidential basis from which the answer-relevant causal evidence $C$ and the statistically correlated but non-causal factors $N$ can be disentangled.

\begin{figure}[t]
    \centering
    \begin{subfigure}{0.305\linewidth}
        \centering
        \includegraphics[width=\linewidth]{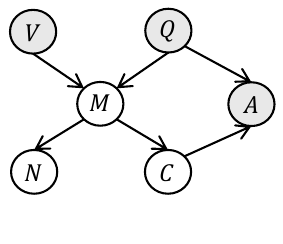}
        \vspace{-15pt}
        \caption{}
        \label{fig:causality_videoqa}
    \end{subfigure}
    \hspace{1mm}
    \begin{subfigure}{0.305\linewidth}
        \centering
        \includegraphics[width=\linewidth]{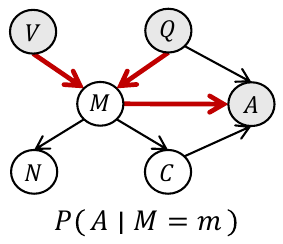}
        \vspace{-15pt}
        \caption{}
        \label{fig:causality_generic_videoqa}
    \end{subfigure}
    \hspace{1mm}
    \begin{subfigure}{0.305\linewidth}
        \centering
        \includegraphics[width=\linewidth]{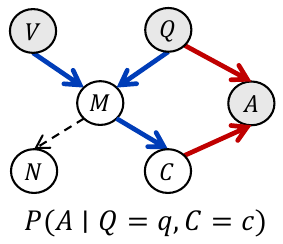}
        \vspace{-15pt}
        \caption{}
        \label{fig:causality_genuine_videoqa}
    \end{subfigure}
    \vspace{-2pt}
    \caption{Structural causal abstraction of VideoQA reasoning.}
    \label{fig:causality_videoqa}
    \vspace{-14pt}
\end{figure}

As shown by the red edges in Fig.~\ref{fig:causality_generic_videoqa}, mainstream VideoQA models typically encode and integrate vision-language inputs and predict answers based on the cross-modality representation, i.e., $P(A \mid M=m), \ m=M(V=v,Q=q)$.
However, conventional paradigms typically capture all statistical patterns including the correlations between $M$ and $C$ or $N$.
As such, the non-causal path $N \leftarrow M \rightarrow A$ forms a common-cause structure, implying that although $N$ and $A$ are not causally related, they exhibit statistical dependence ($N \not\perp A$). Consequently, the reasoning process can be confounded by non-causal factors from $N$.

Eliminating non-causal dependencies through the D-separation criterion provides a practical solution: by controlling the representation $M$, all non-causal paths between $N$ and $A$ are blocked, thereby encouraging the desired conditional independence $N \perp A \mid M$.
As shown in Fig.~\ref{fig:causality_genuine_videoqa}, the desired causal reasoning process blocks the influence of $N$ by identifying $C=c$ from $M$ (i.e., the blue edges) and preserves the causal path $Q,C \rightarrow A$ (i.e., the red edges), which can be formulated as $P(A \mid Q=q, C=c, N=n)=P(A \mid Q=q, C=c)$, reflecting the underlying mechanism of robust reasoning.

To emulate such robust reasoning, we propose a causal intervention–based learning framework. Within this framework, we conduct cross-modality disentanglement that explicitly identifies the causal and non-causal visual representations, $F_c$ and $F_{nc}$. The model then performs inference under the conditions $C = F_c$ and $Q = q$, i.e., $P(A \mid Q = q, C = F_c)$. Note that the proposed SCM is not intended to recover the complete data-generating mechanism. In practice, our objective is to learn a representation $M$ (and its disentangled components $C$ and $N$) that approximately satisfies the intended conditional independencies, rather than to assert full knowledge of the true data-generating mechanism.

\section{The Proposed Method}

\begin{figure*}[t]
    \centering
    \includegraphics[width=0.960\textwidth]{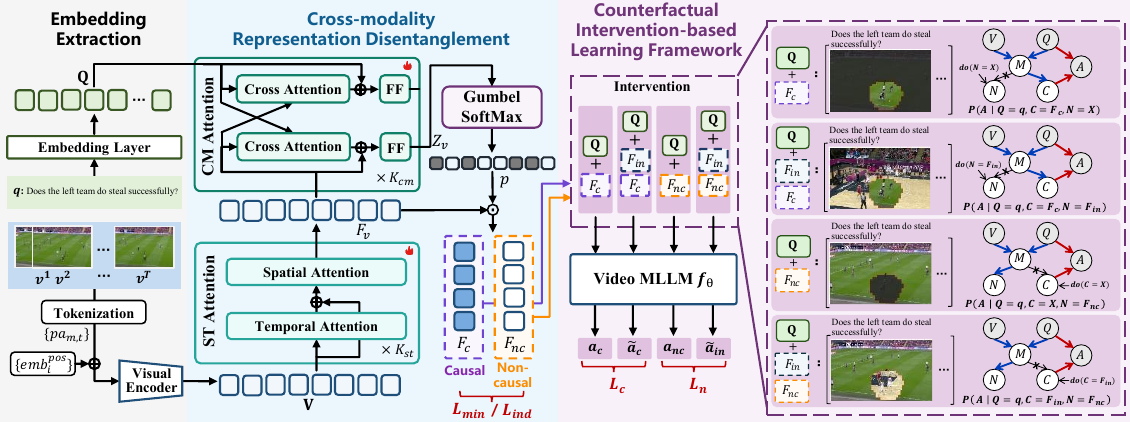}
    \caption{Overview of the proposed CREDiT framework.}
    \label{fig:overall framework}
    \vspace{-13pt}
\end{figure*}  
As illustrated in Fig.~\ref{fig:overall framework}, the proposed CREDiT framework comprises three main components: embedding extraction, cross-modality representation disentanglement, and a counterfactual intervention-based learning framework.

\subsection{Tokenization}
The embedding extraction process is similar to that of common video MLLMs~\cite{lin2023videollava,bai2023qwenvl}. For an input video consisting of $T$ frames, each frame is first tokenized into $M$ patches. A pretrained video encoder is then applied to obtain the visual embeddings
$\mathbf{V} = \left[ v_{{1{:}M,\,1{:}T}} \right] \in \mathbb{R}^{MT \times D},$
where $v_{m,t}$ denotes the embedded feature representation corresponding to the $m$-th image patch of the $t$-th frame, i.e., $pa_{m,t}$, and $D$ is the embedding dimension.
Similarly, a question $q$ of length $N$ is transformed into a sequence of text embeddings
$ \mathbf{Q} = \left[ w_{{1{:}N}}\right] \in \mathbb{R}^{N \times D},$
where $w_n$ represents the embedding of the $n$-th token.
While mainstream MLLMs directly perform reasoning using $\mathbf{V}$ and $\mathbf{Q}$, CREDiT operates at a deeper structural level by exploiting them for causal disentanglement.

\subsection{Cross-modality Disentanglement}
In the cross-modality representation causal disentanglement stage of CREDiT, two learnable Transformer-based networks~\cite{vaswani2017transformer}, namely spatiotemporal (ST) attention and cross-modality (CM) attention, are employed to capture higher-order contextual dependencies. Based on the fused representations, independence and minimality constraints are further imposed to facilitate causal and non-causal disentanglement.

\textbf{Spatiotemporal and Cross-modality Attention.}
Due to large variations in the spatiotemporal scales of question-related visual evidence in real-world scenarios, we follow the design in~\cite{bertasius2021spacestattention} and adopt a spatiotemporal attention network $\Phi_{st}(\cdot)$ with $K_{st}$ stacked layers, using $\mathbf{V}$ as the initial input.
Each layer sequentially performs temporal attention across frames and spatial attention across patches, enabling each visual token to aggregate motion-related and region-level contextual information.
After the $K_{st}$ layers, the video representation is obtained as
\begin{equation}\label{Eq:output_of_stattention}
F_v = [f_{1{:}M,1{:}T}] = \Phi_{st}(\mathbf{V}),
\end{equation}
where $F_v$ denotes the encoded visual features, and $M$ and $T$ denote the number of patches and frames, respectively.

After video modeling, we employ a cross-modality attention network $\Phi_{cm}(\cdot)$ with $K_{cm}$ stacked layers to capture vision-language correlations, taking $\mathbf{F}_v$ and $\mathbf{Q}$ as inputs.
Each layer performs bidirectional interaction between visual and textual tokens, where visual features attend to question representations and textual features attend to video representations.
The cross-modality encoded outputs are then obtained as
\begin{equation}\label{Eq:cm_output}
Z_v, Z_q = \Phi_{cm}(F_v, \mathbf{Q}),
\end{equation}
where $Z_v$ and $Z_q$ are cross-modality representations corresponding to $F_v$ and $\mathbf{Q}$, respectively.
The detailed computations of the temporal, spatial, and cross-modality attention layers are presented in the Supplementary Materials.


\textbf{Causal Disentanglement and Constraints.}
After cross-modality fusion, our method generates discriminative indicators based on $Z_v$ to determine which features in $F_v$ are causal. Considering that soft masking methods such as Softmax may retain correlated non-causal features~\cite{arjovsky2019invariant}, discrete feature selection is more conducive to achieving sharper disentanglement. Since discrete operations are generally non-differentiable, we adopt Gumbel-Softmax~\cite{Jang_Gu_Poole_2016GumbelSoftmax} to approximate discrete sampling while preserving differentiability for gradient-based optimization. Therefore, the discrete discriminative indicators $\{ p_{{1{:}M,1{:}T}} \}$ are generated as:
\begin{equation}
\label{discriminative_label}
\{ p_{{1{:}M,1{:}T}} \}
= \text{Gumbel-Softmax}(\text{MLP}(Z_v)),
\end{equation}
where each vector $p_{m,t} \in \mathbb{R}^{2}$ indicates whether the feature $f_{m,t}$ and its corresponding patch $pa_{m,t}$ are causally relevant to the reasoning process. 
Specifically, $p_{m,t}[0] = 1, p_{m,t}[1] = 0$ indicates causal relevance, whereas $p_{m,t}[0] = 0, p_{m,t}[1] = 1$ indicates non-causal relevance.

Utilizing the discriminative indicators, the method disentangles the visual features into causal and non-causal sets:
\begin{equation}
\begin{aligned}
I_c = \{(m,t) \mid p_{m,t}[0] = 1\},
I_n = \{(m,t) \mid p_{m,t}[1] = 1\},
\end{aligned}
\end{equation}
\begin{equation}
\begin{aligned}
F_c &= \left[\{f_{m_i,t_i}\}_{(m_i,t_i)\in I_c,\ i=1,\ldots,|I_c|}\right],\\
F_{nc} &= \left[\{f_{m_i,t_i}\}_{(m_i,t_i)\in I_n,\ i=1,\ldots,|I_n|}\right].
\end{aligned}
\end{equation}

Although the visual representations are explicitly separated, semantic entanglement may still persist. Specifically, non-causal signals may inadvertently leak into ${F}_{c}$, violating the desired conditional independence $N \perp C \mid M$.

To further enforce independence between ${F}_{c}$ and ${F}_{nc}$, we introduce the Hilbert-Schmidt Independence Criterion (HSIC)~\cite{gretton2005measuring}, a non-parametric kernel-based measure that can capture higher-order statistical dependencies between random variables.
Such independence penalty is defined as:
\begin{equation}\label{eq:ind}
    \mathcal{L}_{\text{ind}} = \text{HSIC}({F}_{c}, {F}_{nc}).
\end{equation}

Additionally, to prevent ${F}_{c}$ from carrying redundant or irrelevant context and weakening the disentanglement, we impose a minimality constraint by applying a sparsity regularizer on the discriminative indicators:
\begin{equation}\label{eq:min}
    \mathcal{L}_{\text{min}} = \sum_{m=1}^{M} \sum_{t=1}^{T} p_{m,t}[0].
\end{equation}


\subsection{Counterfactual Intervention-based Learning}
To facilitate the cross-modality disentanglement module's ability to distinguish causal representations, we propose a counterfactual learning framework that integrates causal interventions and leverages a pretrained MLLM decoder.

According to causal theory, causal interventions conceptually sever the dependence between a variable and its parent nodes~\cite{pearl2009causal}. Inspired by this principle, our method operates in the representation space and constructs counterfactual configurations to approximate the intended intervention effects under the proposed SCM, enabling the capture of answer-relevant information while decoupling non-causal associations, as illustrated by the third block in Fig.~\ref{fig:overall framework}.
Since the position-aware structure of Transformers preserves the token-level correspondence between $\left[pa_{1{:}M,1{:}T}\right]$ and the features in $F_v$ and $Z_v$, each feature $f_{m,t}$ can serve as a proxy for $pa_{m,t}$. Therefore, interventions are performed at the feature level rather than on raw frames.

\textbf{Causal Information Capture.}
To ensure that $F_c$ retains sufficient causal information for predicting the correct answer, we replace the original non-causal representation $F_{nc}$ with random noise $X \sim \mathcal{N}(0,1)$ in the first input configuration, which can be viewed as the intervention $do(N = X)$. Under this intervention, the edge inducing spurious correlations is blocked ($N \not\leftarrow M$), while the answer-relevant causal pathway $C \rightarrow A$ is preserved. The predicted answer distribution of $f_{\theta}$ becomes
\begin{equation}
a_{c} = P_{\theta}(A \mid Q = q, C = F_{c}, N = X).
\end{equation}

Building on the first input, $N = X$ or other non-causal factors may still exhibit statistical dependence on the prediction. To encourage the invariance of $C \rightarrow A$ under arbitrary non-causal perturbations, the second input applies another intervention on $N$, i.e., $do(N = F_{in})$, where $F_{in}$ is the representation of a randomly selected sample from the dataset. The corresponding inference process is given by
\begin{equation}
\tilde{a}_{c} = P_{\theta}(A \mid Q = q, C = F_{c}, N = F_{in}).
\end{equation}

In these two input configurations, interventions on $N$ block non-causal correlations while preserving the causal path containing $F_c$, thereby enforcing the sufficiency and robustness of the causal information encoded in $F_c$. Consequently, the answering process is encouraged to follow the desired causal mechanism. Therefore, the training objective should encourage the answer distributions produced from the causal evidence, namely $a_c$ and $\tilde a_c$, to approach the ground-truth answer distribution $a_{gt}$.

\textbf{Non-causal Dependency Isolation.}
Beyond modeling the causal relationship, the model is expected to capture the non-causal effect of $N$ that should be represented in $F_{nc}$.
Once non-causal clues are captured and isolated, residual spurious correlations can be explicitly removed from $F_c$. To this end, the third input retains $F_{nc}$ while applying the intervention $do(C = X)$, which removes paths involving $C$. In this case, causal information is eliminated, and the output distribution reflects the spurious correlations originating from $N$:
\begin{equation}
a_{nc} = P_{\theta}(A \mid Q = q, C = X, N = F_{nc}) .
\end{equation}

Similarly, to prevent $N$ from forming a non-causal dependency with $A$ through possible visual cues, the fourth input replaces $F_{c}$ with a randomly selected $F_{in}$ as another intervention on $C$, thereby further ensuring the ineffectiveness of the non-causal signals encoded in $F_{nc}$ for answer reasoning. The corresponding output distribution becomes
\begin{equation}
\tilde{a}_{in} = P_{\theta}(A \mid Q = q, C = F_{in}, N = F_{nc}) .
\end{equation}

Under these interventions on $C$ that block causal clues, the model can assess the association between $N$ and $A$. Under the desired SCM mechanism, $N$ should not provide answer-discriminative information once $C$ is removed or controlled. Accordingly, the non-causal outputs (i.e., $a_{nc} $ and $ \tilde{a}_{in}$) revert to a blind guess without any visual prior, and are encouraged to approximate the text-only prediction $P_{\theta}(A \mid Q = q)$.

\subsection{Model Optimization}

The optimization objective of CREDiT consists of five components.
First, a generic VideoQA training objective, $\mathcal{L}_{\text{vqa}}$, is introduced. Since discrete sampling via Gumbel-Softmax may cause training instability and randomness, particularly in the early training stages, this objective provides a representation anchor to stabilize the learning and preserve the quality of global features. Specifically, the complete $F_v$ is fed into $f_{\theta}$ to obtain the prediction $\hat{a}$ and compute the loss:
\begin{equation}
\begin{aligned}
\mathcal{L}_{\text{vqa}} = {\rm CELoss}(a_{gt}, \hat{a}),
\end{aligned}
\end{equation}
where CELoss refers to the cross-entropy loss.

The second term, $\mathcal{L}_{\text{c}}$, encourages the learned $F_{c}$ to encode sufficient causal visual content for answering:
\begin{equation}
\begin{aligned}
\mathcal{L}_{\text{c}} = {\rm CELoss}(a_{gt}, a_c) + {\rm CELoss}(a_{gt}, \tilde{a}_{c}).
\end{aligned}
\end{equation}

The third term, $\mathcal{L}_{\text{n}}$, enforces the ineffectiveness of $F_{nc}$ in the reasoning process, thereby insulating non-causal correlations. Specifically, $a_{nc}$ and $\tilde{a}_{in}$ are encouraged to collapse to the prediction of the text-only input $f_{\theta}(X, q)$, where the visual content is replaced with random noise $X$:
\begin{equation}\label{equation:noncausal}
\mathcal{L}_{\text{n}} = {\rm KL}(f_\theta(X, q) \parallel a_{nc}) + {\rm KL}(f_\theta(X, q) \parallel \tilde{a}_{in}),
\end{equation}
where KL denotes the Kullback--Leibler divergence. The text-only prediction is treated as a target distribution during this regularization.

Together with the disentanglement constraints in Equations~\eqref{eq:ind} and~\eqref{eq:min}, the total loss of CREDiT is formulated as
\begin{equation}
\mathcal{L}_{\text{total}}
= \mathcal{L}_{\text{vqa}}
+ \lambda_{\text{c}}\mathcal{L}_{\text{c}}
+ \lambda_{\text{n}} \mathcal{L}_{\text{n}}
+ \lambda_{\text{ind}} \mathcal{L}_{\text{ind}}
+ \lambda_{\text{min}} \mathcal{L}_{\text{min}},
\end{equation}
where $\lambda_{\text{c}}$, $\lambda_{\text{n}}$, $\lambda_{\text{ind}}$, and $\lambda_{\text{min}}$ are hyperparameters balancing the contributions of different loss components.
During inference, CREDiT leverages the disentangled causal visual representation $F_{c}$ and text embeddings as inputs to the MLLM decoder for answer prediction.

\section{Experiments}
In this section, we conduct comprehensive experiments to evaluate the effectiveness of the proposed CREDiT framework. Specifically, our experiments are designed to answer the following research questions:

\textbf{RQ1: Does CREDiT improve VideoQA performance?} 
The model is evaluated on NExT-GQA~\cite{xiao2024nextgqa} to examine its QA accuracy (Acc@QA) in generic situations. Moreover, recent evidence~\cite{li2024sportsqadatset,yang2024sportsIntelligence} suggests that MLLMs exhibit a performance gap in sports scenarios due to the high variability in spatiotemporal scales and diverse backgrounds. Consequently, extensive experiments are conducted on Sports-QA~\cite{li2024sportsqadatset} and SPORTU-video~\cite{xia2024sportudataset} to further validate CREDiT's efficacy in challenging realistic contexts.

\textbf{RQ2: Can CREDiT improve trustworthy reasoning by identifying causal evidence?} 
The ability to identify causal video content via the proposed disentanglement module, as well as its impact on QA performance, is evaluated using intersection over prediction/union (IoP/U) and grounded QA accuracy (Acc@GQA) on NExT-GQA.

\textbf{RQ3: Are the proposed components effective and well-behaved?} 
Ablation studies and hyperparameter analyses are conducted to examine the contribution and sensitivity of each component.

\textbf{RQ4: Does CREDiT effectively disentangle causally relevant evidence?}
Quantitative and qualitative analyses are performed to verify whether the disentangled causal representations capture answer-relevant evidence while suppressing non-causal associations.

\begin{table*}[t]
\centering
\small 
\begin{minipage}[t]{0.53\textwidth}
    \centering
    \caption{QA Accuracy (\%) of different models on the Sports-QA dataset. The results for VideoChat are obtained via our re-implementation, while others are directly cited from existing research~\cite{li2024sportsqadatset,yang2024sportsIntelligence,kim2024videoicl}.}
    \label{table:accuracy_different_models_sports_qa}
    \vspace{6pt}
    \begin{NiceTabular}{lccccc}[colortbl-like]
    \toprule
    Model & {Overall} & {Des.} & {Tem.} & {Cau.} & {Cou.} \\
    \midrule
    CoMem~\cite{gao2018motionappearance} & 43.7 & 51.7 & 32.7 & 48.7 & 51.3 \\
    HME~\cite{fan2019HMEheterogeneous} & 57.1 & 77.0 & 32.9 & 50.0 & 60.9 \\
    HGA~\cite{jiang2020reasoninghga} & 58.0 & 76.5 & 31.2 & 50.4 & 54.3 \\
    IGV~\cite{li2022invariantgroundingvqa} & 58.3 & 78.1 & 34.4 & 52.0 & 58.2 \\
    HQGA~\cite{xiao2022videohqga} & 58.9 & 77.6 & 33.8 & 54.1 & 58.2 \\
    MASN~\cite{seo2021attendmasn} & 57.0 & 78.4 & 33.6 & 52.2 & 58.2 \\
    AFT$_{\text{GloVe}}$~\cite{li2024sportsqadatset} & 59.2 & 78.9 & 35.3 & 55.1 & 56.3 \\
    \midrule
    LLaVA-Video-7B~\cite{zhang2024llavavideonew} & 25.5 & 39.1 & 12.0 & 22.9 & 28.2 \\
    \rowcolor{lightgreen}Qwen2VL-7B~\cite{wang2024qwen2vl} & 26.8 & 42.3 & 14.6 & 23.7 & 27.0 \\
    \rowcolor{lightgreen}Qwen2.5VL-7B~\cite{2025Qwen2.5vl} & 28.4 & 44.1 & 16.0 & 25.2 & 27.9 \\
    \rowcolor{lightgreen}Video-LLaVA-7B~\cite{lin2023videollava} & 30.3 & 42.8 & 13.5 & 36.8 & 42.7 \\
    VideoICL$_{\text{Qwen2VL-7B}}$~\cite{kim2024videoicl} & 51.5$_{\Delta \text{23.7}}$ & - & - & - & - \\
    VideoChat-R1.5$^\dagger$$^\ddagger$~\cite{kim2024videoicl} & 49.1$_{\Delta \text{20.7}}$ & 71.3 & 27.6 & 43.1 & 44.0 \\
    \midrule
    \rowcolor{lightred} 
    CREDiT$_{\text{Qwen2VL-7B}}$ & 57.4$_{\Delta \text{30.6}}$ & 74.9 & 32.8 & 51.3 & 56.1 \\
    \rowcolor{lightred} 
    CREDiT$_{\text{Qwen2.5VL-7B}}$ & 59.6$_{\mathbf{\Delta} \textbf{31.2}}$ & 78.2 & \textbf{35.5} & 54.3 & 58.6 \\
    \rowcolor{lightred} 
    CREDiT$_{\text{Video-LLaVA-7B}}$ & \textbf{60.4}$_{\Delta \text{30.1}}$ & \textbf{79.4} & 34.7 & \textbf{55.4} & \textbf{61.2} \\
    \bottomrule
    \end{NiceTabular}
\end{minipage}
\hfill 
%
\begin{minipage}[t]{0.42\textwidth}
    \centering
    \caption{QA Accuracy (\%) of different models on SPORTU-video. The results of existing methods are cited from the study~\cite{xia2024sportudataset}.}
    \label{table:Accuracy_of_different_models_on_SPORTU_video_dataset}
    \small
    \begin{NiceTabular}{llc}[colortbl-like] 
        \toprule
        Model & Parameters & Acc@QA \\
        \midrule
        \multicolumn{3}{l}{Closed-source Models} \\
        \midrule
        Claude-3.0-Haiku~\cite{anthropic2024claude3} & - & 47.9 \\
        Claude-3.5-Sonnet~\cite{anthropic2024claude3} & 175B & 70.1 \\
        Gemini 1.5 Pro~\cite{team2024gemini15} & - & 64.9 \\
        Gemini 1.5 Flash~\cite{team2024gemini15} & 8B & 62.5 \\
        GPT-4o mini~\cite{openai2024hellogpt4o} & 8B & 58.1 \\
        GPT-4o~\cite{openai2024hellogpt4o} & 200B & 68.8 \\
        \midrule
        \multicolumn{3}{l}{Open-source Models} \\
        \midrule
        LLaVA-NeXT~\cite{liu2024llavanext} & 7B & 63.7 \\
        Tarsier~\cite{wang2024tarsier} & 7B & 60.1 \\
        Video-ChatGPT~\cite{maaz2023videochatgpt} & 7B & 34.0 \\
        VideoChat2~\cite{li2024mvbenchVideoChat2} & 7B & 61.5 \\
        \rowcolor{lightgreen} Qwen2VL-7B~\cite{wang2024qwen2vl} & 7B & 62.9 \\
        \rowcolor{lightgreen} Qwen2.5VL-7B~\cite{2025Qwen2.5vl} & 7B & 64.2 \\
        \midrule
        \rowcolor{lightred} CREDiT$_{\text{Qwen2VL-7B}}$ & 7B & 70.8$_{{\Delta} 7.9}$ \\
        \rowcolor{lightred} CREDiT$_{\text{Qwen2.5VL-7B}}$ & 7B& \textbf{71.9}$_{\mathbf{{\Delta}} \textbf{7.7}}$\\
        \bottomrule
    \end{NiceTabular}
\end{minipage}
\hspace*{0.018\textwidth}
\vspace{-11pt}
\end{table*}

\subsection{Datasets, Metrics, and Baselines}

\textbf{Datasets.} Experiments are conducted on three datasets: NExT-GQA~\cite{xiao2021nextqadataset}, Sports-QA, and SPORTU-video.
NExT-GQA extends the generic VideoQA dataset NExT-QA by incorporating visual evidence annotations, requiring models to answer questions and ground their reasoning within specific video intervals.
Sports-QA is a large-scale VideoQA benchmark for sports scenarios, covering multiple sports categories and different question types, which requires fine-grained action understanding and long-range temporal reasoning.
SPORTU-video contains slow-motion sports videos and multi-choice QA pairs across multiple sports categories, with questions spanning from action recognition to rule understanding.

\textbf{Metrics.}
Following the setting in~\cite{xiao2024nextgqa}, we employ four evaluation metrics: Acc@QA, IoP, IoU, and Acc@GQA.
Notably, existing public benchmarks only provide grounded QA annotations at the time-interval level, such as NExT-GQA. Therefore, to align CREDiT's patch-level predictions with IoP/U evaluations on NExT-GQA, we convert the predictions into frame-level outputs.
Detailed definitions of the metrics and the patch-to-frame conversion are provided in the Supplementary Materials.

\textbf{Baselines.} 
Diverse VideoQA methods are selected as baselines for a fair and comprehensive evaluation. These methods are broadly categorized into three groups: (1) Classical VideoQA models (e.g., CoMem~\cite{gao2018motionappearance}, HME~\cite{fan2019HMEheterogeneous}, HGA~\cite{jiang2020reasoninghga}, MASN~\cite{seo2021attendmasn}, ATF~\cite{li2024sportsqadatset}), representing early frameworks based on memory or attention mechanisms. (2) MLLMs and their derivatives (e.g., Tarsier~\cite{wang2024tarsier}, Claude series~\cite{anthropic2024claude3}, Gemini series~\cite{team2024gemini15}), which currently achieve SOTA performance through large-scale pretraining. (3) Grounding-aware methods (e.g., IGV~\cite{li2022invariantgroundingvqa}, NG+~\cite{xiao2024nextgqa}, CRA~\cite{chen2025crosscra}, LeAdQA~\cite{dong2025leadqa}, LLoVi~\cite{zhang2024simplellovi}, TOGA~\cite{gupta2025toga}, VideoChat~\cite{li2025videochatR1,NEURIPS2025_VideoChatR1.5}), which explicitly model temporal evidence and serve as the most relevant prior art. 
The green-shaded entries indicate the baseline models adopted by CREDiT.
“$\dagger$” indicates models using additional customized pretraining (e.g., vision-language alignment or temporal grounding).
“$\ddagger$” indicates models trained with ground-truth temporal interval labels.

\subsection{Implementation}
CREDiT is built upon MLLM baselines, Video-LLaVA~\cite{lin2023videollava} and the Qwen-VL series~\cite{wang2024qwen2vl, 2025Qwen2.5vl}. The embedding extraction pipeline, including partitioning, tokenization, and encoding, as well as the generative model $f_\theta(\cdot)$, are constructed following the implementation of the corresponding baseline MLLMs with parameters kept frozen.
The number of layers in the ST module $K_{st}$ and the CM module $K_{cm}$ are set to $4$ and $6$, respectively. The disentangled features are projected to the required hidden dimension of each MLLM decoder. To determine the final answer, we employ a sequence scoring strategy, and the candidate with the highest sequence score is selected as the final prediction.
All experiments are conducted on RTX A6000 GPUs. The detailed architectural settings, optimization hyper-parameters, and training schedules are provided in the Supplementary Material.

\begin{table}[t] %
\centering
\small %

\caption{QA Accuracy (\%) of different methods on NExT-GQA. Tem. and Cau. indicate the temporal and causal question types respectively. The results for VideoChat on Tem. and Cau. are obtained via our re-implementation.}
\label{tab:accqa_on_nextgqa}
\small
\begin{NiceTabular}{lccc}[colortbl-like]
\toprule
Method & Overall & Tem. & Cau. \\
\midrule
{\color{gray!81!black} SeViLA$^\dagger$} & {\color{gray!81!black} 68.1} & {\color{gray!81!black}-} & {\color{gray!81!black}-} \\
CRA$_{\text{FrozenBiLM}}$ & 70.2 & - & - \\
\rowcolor{lightgreen} Qwen2.5VL-7B  & 66.1 & 64.4 & 67.3 \\
{\color{gray!81!black} VideoChat-R1$^\dagger$$^\ddagger$} &
{\color{gray!81!black}\textbf{70.6}$_{\Delta \textbf{4.5}}$} &
{\color{gray!81!black}\textbf{69.0}$_{\Delta \textbf{4.6}}$} &
{\color{gray!81!black}71.6$_{\Delta 4.3}$} \\
{\color{gray!81!black} LeAdQA$_{\text{Qwen2.5VL-7B}}$$^\ddagger$} &
{\color{gray!81!black}67.7$_{\Delta 1.6}$} &
{\color{gray!81!black}65.6$_{\Delta 1.1}$} &
{\color{gray!81!black}69.1$_{\Delta 1.8}$} \\
\midrule
\rowcolor{lightred} 
\rowcolor{lightred} 
CREDiT$_{\text{Qwen2.5VL-7B}}$  & 70.4$_{\mathbf{\Delta}  4.3}$ & 68.5$_{\mathbf{\Delta} 4.1}$ & \textbf{71.7}$_{\mathbf{\Delta} \textbf{4.4}}$ \\
\bottomrule
\end{NiceTabular}

\vspace{0.4cm}

\addtolength{\tabcolsep}{-3.75pt}
\centering
\caption{Performance comparison under the Grounded QA setting on NExT-GQA. }
\label{tab:accGqa_on_nextgqa}
\small
\begin{NiceTabular}{lccccc}[colortbl-like]
\toprule
Method & Acc@GQA & mIoP & IoP@0.5 & mIoU \\
\midrule

\rowcolor{verylightgray}
Human & 82.1 & 72.1 & 86.2 & 61.2 \\

\rowcolor{verylightgray}
Random & 1.7 & 21.1 & 8.7 & 21.1 \\
\midrule

{\color{gray!81!black} SeViLA$^\dagger$~\cite{yu2023SeViLAself}} &
{\color{gray!81!black}16.6} &
{\color{gray!81!black}29.5} &
{\color{gray!81!black}22.9} &
{\color{gray!81!black}21.7} \\


{\color{gray!81!black} VideoStreaming$^\dagger$~\cite{qian2024videostreaming}} &
{\color{gray!81!black}17.8} &
{\color{gray!81!black}32.2} &
{\color{gray!81!black}31.0} &
{\color{gray!81!black}19.3} \\

{\color{gray!81!black} LeAdQA-7B$^\ddagger$~\cite{dong2025leadqa}} &
\color{gray!81!black}19.2 &
\color{gray!81!black}30.3 &
\color{gray!81!black}29.5 &
\color{gray!81!black}20.5 \\

{\color{gray!81!black} Grounded-VideoLLM$^\dagger$$^\ddagger$~\cite{wang2024groundedvideollm}} &
{\color{gray!81!black}26.7} &
{\color{gray!81!black}34.5} &
{\color{gray!81!black}34.4} &
{\color{gray!81!black}21.1} \\




{\color{gray!81!black} VideoChat-R1.5$^\dagger$$^\ddagger$~\cite{NEURIPS2025_VideoChatR1.5}} &
{\color{gray!81!black}61.9} &
{\color{gray!81!black}74.9} &
{\color{gray!81!black}77.9} &
{\color{gray!81!black}-} \\


\midrule

IGV~\cite{li2022invariantgroundingvqa} & 10.2 & 21.4 & 18.9 & 14.0 \\
VGT[RBT]~\cite{xiao2022videovgt} & 14.4 & 25.3 & 25.3 & 3.0 \\
VIOLETv2~\cite{fu2023empiricalvioletv2} & 12.8 & 23.6 & 23.3 & 3.1 \\
NG+$_{\text{FrozenBiLM}}$~\cite{xiao2024nextgqa} & 17.5 & 24.2 & 23.7 & 9.6 \\
TimeCraft$_{\text{FrozenBiLM}}$~\cite{liu2024timecraft} & 18.5 & 26.3 & 24.9 & 13.2 \\
CRA$_{\text{FrozenBiLM}}$~\cite{chen2025crosscra} & 18.8 & 26.5 & 25.9 & 13.5 \\
LLoVi~\cite{zhang2024simplellovi} & 24.3 & 37.3 & 36.9 & 20 \\
TOGA~\cite{gupta2025toga} & 24.6 & 40.5 & 40.6 & \textbf{24.4} \\

\midrule

\rowcolor{lightred}
CREDiT & \textbf{27.9} & \textbf{41.4} & \textbf{41.1} & {20.1} \\

\bottomrule
\end{NiceTabular}
\addtolength{\tabcolsep}{3.75pt}
\vspace{-11pt}
\end{table}

\subsection{Comparison with the SOTA Methods}\label{section:comparison}

To comprehensively evaluate the efficacy of the proposed CREDiT framework, we conduct extensive experiments across three benchmarks under both open-ended and multiple-choice QA paradigms. These datasets cover scenarios ranging from general VideoQA contexts to complex, real-world environments, thereby demonstrating the effectiveness of our approach across diverse VideoQA scenarios.
The results are reported in Tables~\ref{table:accuracy_different_models_sports_qa}, \ref{table:Accuracy_of_different_models_on_SPORTU_video_dataset}, and \ref{tab:accqa_on_nextgqa}, leading to the following findings.

\subsubsection{Limitations of Existing Methods in Sports Scenarios}
A key observation from the empirical results is that Multi-modal Large Language Models (MLLMs) suffer from noticeable degradation when deployed in sports-specific contexts. As shown in Table~\ref{table:accuracy_different_models_sports_qa}, the strong MLLM-based baseline, VideoChat-R1.5, only achieves an overall accuracy of $49.1\%$ on Sports-QA, which underperforms several task-specific non-pretrained classical methods. Similarly, as shown in Table~\ref{table:Accuracy_of_different_models_on_SPORTU_video_dataset}, large-scale models such as Claude-3.5 and GPT-4o reach at most $70.1\%$ accuracy on SPORTU-video. This observation is consistent with prior findings~\cite{yang2024sportsIntelligence,Chaudhari_2024}, suggesting that generic pretraining alone is insufficient for robust reasoning in sports scenarios. Such limitations can be attributed to the motion complexity of sports scenes, where the required visual evidence varies drastically across space and time.

\subsubsection{Performance of CREDiT in Complex Sports Contexts}
In contrast, CREDiT achieves the best results among the compared methods across both sports benchmarks. On Sports-QA, $\text{CREDiT}{\text{Video-LLaVA-7B}}$ achieves the best reported result with an overall accuracy of $60.4\%$. On SPORTU-video, $\text{CREDiT}{\text{Qwen2.5VL-7B}}$ achieves a peak accuracy of $71.9\%$, outperforming even Claude-3.5 by $1.8\%$ while using a 7B backbone. These results suggest that explicitly modeling causal relationships and reasoning over disentangled causal visual content enables more reliable reasoning in complex sports contexts.

\subsubsection{Competitive Performance in General Scenarios}
As evidenced by the results in Table~\ref{tab:accqa_on_nextgqa}, VideoChat-R1 achieves the best overall performance ($70.6\%$), benefiting from its multi-task pretraining paradigm and the exploitation of massive NExT-GQA samples. Without relying on customized external multi-task data or annotations, CREDiT yields a comparable overall accuracy of $70.4\%$, outperforming $\text{CRA}_{\text{FrozenBiLM}}$ and narrowing the gap with VideoChat-R1 to only $0.2\%$. Crucially, CREDiT achieves the highest accuracy on causal questions ($71.7\%$) and improves its Qwen2.5VL-7B backbone by $4.3\%$ with minimal training cost.

Across the above benchmarks, the results demonstrate that while large-scale pretraining remains advantageous in generic settings, our counterfactual reasoning paradigm offers a more efficient and scalable approach to diverse video understanding.

\subsection{Validation of the Proposed Methods}

To evaluate the ability to extract visual evidence, we conduct comparative experiments under the grounded QA setting provided by NExT-GQA. The results are shown in Table~\ref{tab:accGqa_on_nextgqa}. Following prior works, existing methods are grouped into approaches with additional training resources and approaches trained solely using ground-truth answers. To align with the interval-level evaluation protocol, patch-level outputs are converted into frame-level predictions.

\subsubsection{Performance of Methods with Additional Training Resources}
Several approaches have demonstrated strong GQA performance when supported by targeted auxiliary pre-training resources. In particular, VideoChat benefits from additional supervision, including temporal annotations from NExT-GQA, resulting in substantially higher Acc@GQA and IoP scores. However, as discussed in Section~\ref{section:comparison}, such training requirements are often unavailable in practical VideoQA scenarios.

\subsubsection{Grounding Capability of CREDiT}
Among methods trained without additional supervision, TOGA represents a strong baseline through deconfounding and grounding-aware optimization. Compared with approaches without additional training assistance, CREDiT achieves the best Acc@GQA and mIoP scores, improving Acc@GQA from $24.6\%$ to $27.9\%$ and mIoP from $40.5\%$ to $41.4\%$. These results indicate that CREDiT can localize answer-relevant evidence more reliably while maintaining strong QA reasoning ability.

Although CREDiT obtains lower IoU performance, the grounding annotations are provided as continuous temporal intervals that may contain frames with weak causal relevance. In contrast, the sparse patch-level evidence identified by CREDiT focuses on more precise answer-related regions, leading to stronger IoP performance.

Overall, the improvements on Acc@GQA and IoP indicate that CREDiT improves answer-relevant evidence localization with modest training cost, leading to more trustworthy grounding and reasoning.

\subsection{Ablation Study}

To evaluate the contributions of different components in CREDiT, we progressively introduce the optimization objectives, including the foundational VideoQA loss $\mathcal{L}_{\text{vqa}}$, causal capture loss $\mathcal{L}_{\text{c}}$, non-causal capture loss $\mathcal{L}_{\text{n}}$, and the disentanglement constraints ($\mathcal{L}_{\text{ind}}+\mathcal{L}_{\text{min}}$). Results are reported in Table~\ref{tab:ablation_detail}.

Adding $\mathcal{L}_{\text{vqa}}$ consistently improves performance over the frozen backbone, providing a stable optimization anchor for the newly introduced modules. Introducing the causal objective $\mathcal{L}_{\text{c}}$ leads to the largest performance gain, boosting Sports-QA from $36.4\%$ to $52.6\%$ and improving Acc@GQA on NExT-GQA from $6.1\%$ to $18.6\%$. This demonstrates the effectiveness of the proposed counterfactual intervention strategy in capturing answer-relevant evidence.

\begin{table}[t]
\addtolength{\tabcolsep}{-3.85pt}
\centering
\caption{Ablation study results. $\mathcal{L}_{\text{vqa}}$, $\mathcal{L}_{\text{c}}$, $\mathcal{L}_{\text{n}}$, 
and the combined disentanglement constraints ($\mathcal{L}_{\text{ind}} +\mathcal{L}_{\text{min}}$) denote different components in the proposed CREDiT framework. The baseline model is the pretrained Qwen2.5VL-7B. For Acc@GQA, $\Delta$ is computed relative to the $\mathcal{L}_{\text{vqa}}$ setting due to the absence of baseline grounding output.}
\label{tab:ablation_detail}
\small
\begin{NiceTabular}{lcccc}[colortbl-like]
\toprule
\multirow{2.5}{*}{Setting} &
\multirow{2.5}{*}{Sports-QA} &
\multirow{2.5}{*}{SPORTU-video} &
\multicolumn{2}{c}{NExT-GQA} \\
\cmidrule(lr){4-5}
& & & Acc@QA & Acc@GQA \\
\midrule
\rowcolor{lightgreen} Baseline
& 28.4 & 64.2 & 66.1 & - \\
\midrule
$+\mathcal{L}_{\text{vqa}}$
& 36.4$_{\Delta8.0\,\,\,}$
& 66.0$_{\Delta 1.8}$
& 67.5$_{\Delta 1.4}$
& 6.1 \\
$+\mathcal{L}_{\text{c}}$
& 52.6$_{\Delta 24.2}$
& 69.4$_{\Delta 5.2}$
& 68.9$_{\Delta 2.8}$
& 18.6$_{\Delta 12.5\,\,\,}$ \\
$+\mathcal{L}_{\text{n}}$
& 55.7$_{\Delta 27.3}$
& 71.6$_{\Delta 7.4}$
& 69.8$_{\Delta 3.7}$
& 24.2$_{\Delta 18.1}$ \\
\rowcolor{lightred} $+\mathcal{L}_{\text{ind}}+\mathcal{L}_{\text{min}}$
& \textbf{59.6}$_{\Delta 31.2}$
& \textbf{71.9}$_{\Delta 7.7}$
& \textbf{70.4}$_{\Delta 4.3}$
& \textbf{27.9}$_{\Delta 21.8}$ \\
\bottomrule
\end{NiceTabular}
\vspace{-4pt}
\addtolength{\tabcolsep}{3.85pt}
\end{table}

\begin{figure}[t]
    \centering
    \scalebox{1}[0.9]{
        \includegraphics[width=0.92\linewidth]{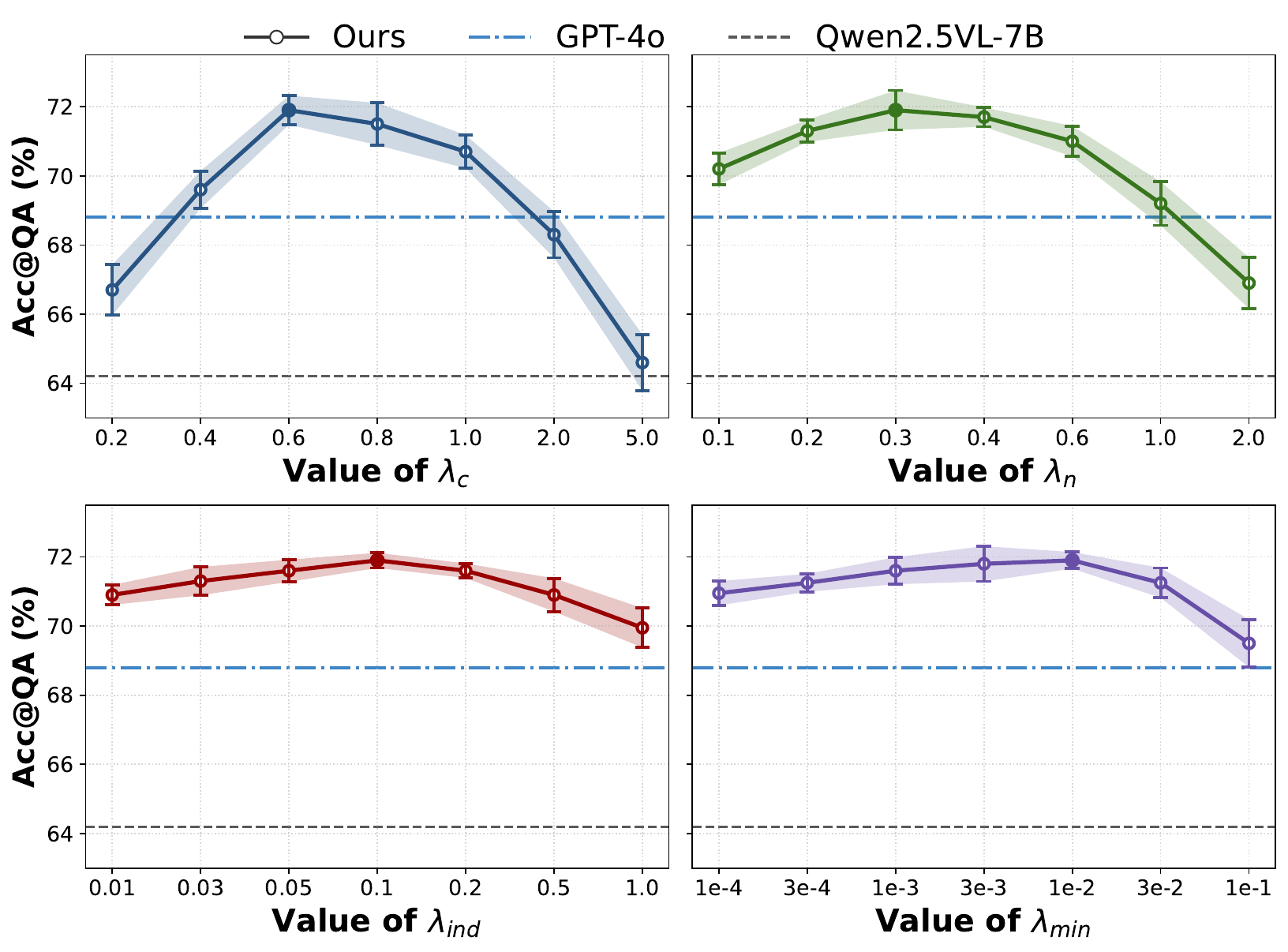}
    }
    \vspace{-4pt}
    \caption{Hyper-parameter sensitivity analysis of four loss weights on SPORTU-video. GPT-4o and Qwen2.5VL-7B are used as baselines. The shaded regions denote the standard deviation over repeated runs.}
    \label{fig:hyper_param}
    \vspace{-11pt}
\end{figure}

Further incorporating $\mathcal{L}_{\text{n}}$ yields consistent improvements across all datasets, indicating that explicitly constraining non-causal information helps reduce the influence of confounding cues. Finally, adding the combined disentanglement constraints completes the full CREDiT framework, which delivers the best performance, including $59.6\%$ on Sports-QA, $71.9\%$ on SPORTU-video, and $27.9\%$ Acc@GQA on NExT-GQA. These results suggest that causal and non-causal modeling are complementary, while the disentanglement constraints further improve the quality of causal evidence separation and reasoning faithfulness.

\subsection{Hyper-parameter Sensitivity Analysis}

To evaluate the stability of CREDiT, we analyze the sensitivity of four loss weights, namely $\lambda_c$, $\lambda_n$, $\lambda_{ind}$, and $\lambda_{min}$. The results on SPORTU-video are shown in Fig.~\ref{fig:hyper_param}. Specifically, the evaluation grid is sampled densely around the expected optimal regions and becomes sparser toward the outer boundaries. To reduce randomness, each experiment is repeated, and the shaded regions denote the standard deviation.

As illustrated in Fig.~\ref{fig:hyper_param}, CREDiT consistently outperforms both Qwen2.5VL-7B and GPT-4o across a broad range of parameter settings.
The best average performance is observed at $\lambda_c = 0.6$ and $\lambda_n = 0.3$ and remains stable with $\lambda_c$ within $[0.4,1.0]$ and $\lambda_n$ within $[0.1,1.0]$.
Similar behavior can be observed for the disentanglement weights, where
the model reaches the optimum when $\lambda_{ind}$ and $\lambda_{min}$ are set to $0.1$ and $10^{-2}$, respectively.
Moreover, the standard deviations remain small across all settings, indicating that the performance variations are not caused by random fluctuations.

Overall, the results demonstrate that CREDiT is robust to hyper-parameter variations and maintains stable performance within a relatively wide parameter range. Similar observations on NExT-GQA are provided in Supplementary Materials.

\subsection{Analysis of Disentangled Visual Evidence}


To verify whether CREDiT successfully disentangles answer-relevant evidence from non-causal context, we conduct both quantitative and qualitative analyses. Specifically, we first evaluate the performance using only the causal $F_c$ or non-causal representation $F_{nc}$. We then visualize the spatial-temporal localization masks generated by CREDiT alongside the baseline Qwen2.5VL-7B to further examine the learned evidence. Since the frozen Qwen backbone does not natively generate grounding outputs, its attention is visualized via post-hoc gradient-based techniques. In contrast, CREDiT projects its patch-level causal indicators through token-to-patch correspondences in ViT-based networks. Additionally, more visualized cases are presented in Supplementary Materials.

\begin{table}[t]
\centering
\caption{Acc@QA with causal $F_c$ or non-causal $F_{nc}$ inputs.}
\label{tab:fc_fnc_analysis}
\begin{tabular}{lccc}
\toprule
Setting & Sports-QA & SPORTU-video & NExT-GQA \\
\midrule
Input $F_{nc}$ & 25.7  & 30.1 & 32.2 \\
Input $F_c$   & \textbf{59.6} & \textbf{71.9} & \textbf{70.4} \\
\bottomrule
\vspace{-18pt}
\end{tabular}
\end{table}

\begin{figure}[t]  
    \centering
    \begin{subfigure}[b]{0.93\linewidth}
        \centering
        \includegraphics[width=\linewidth]{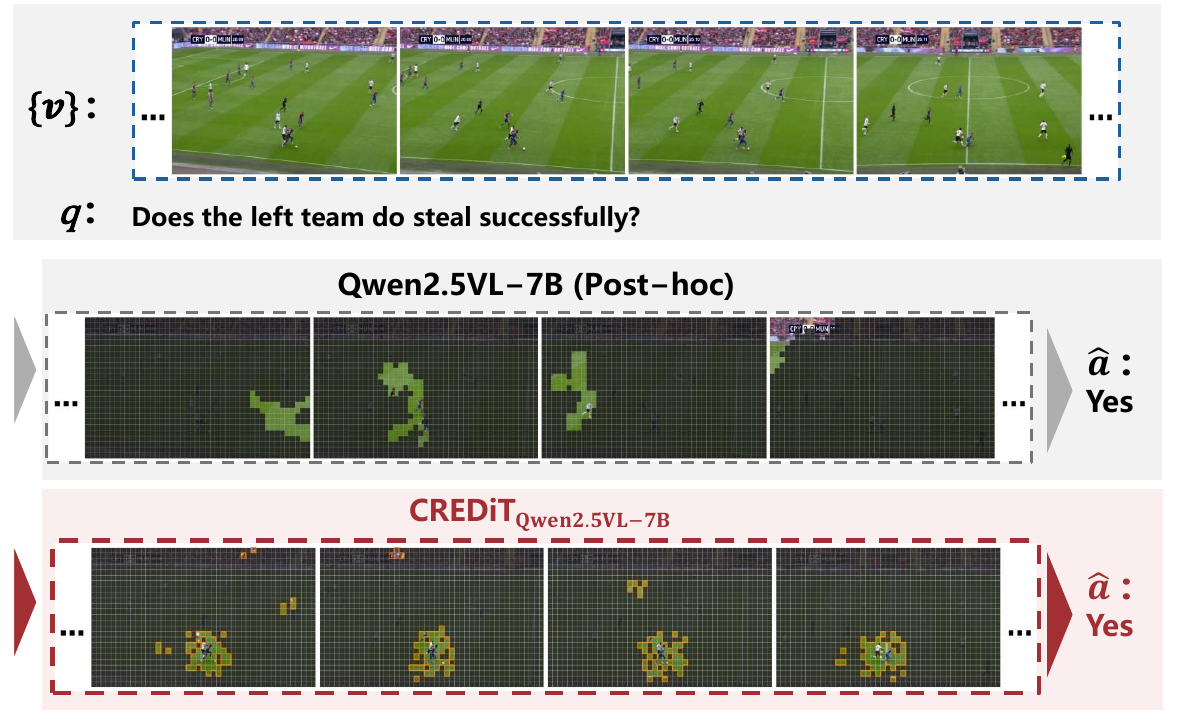}
        \label{fig:fcase1}
    \end{subfigure}
    \vspace{-12pt} 
    \caption{Successful case of CREDiT.}
    \label{fig:all_cases}
    \vspace{-4pt}
\end{figure}

\begin{figure}[t]  
    \centering
    \begin{subfigure}[b]{0.83\linewidth}
        \centering
        \includegraphics[width=\linewidth]{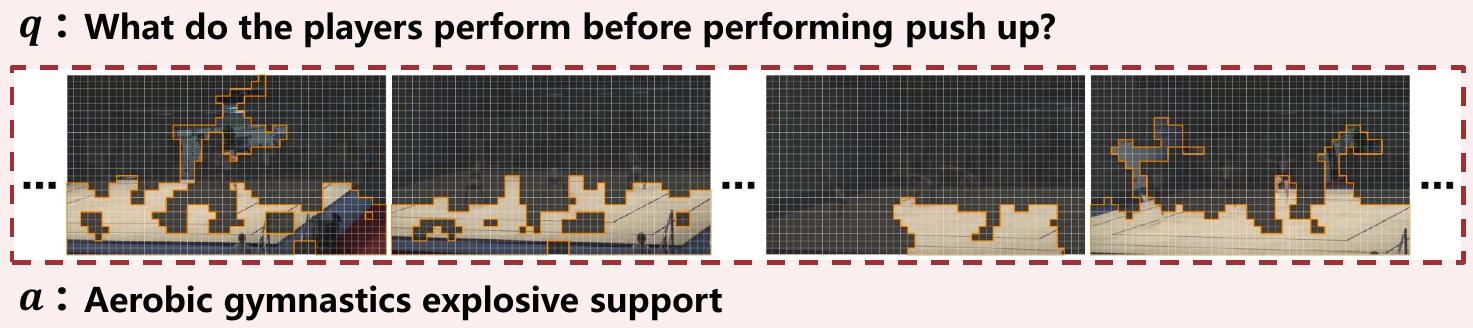}
        \vspace{-13pt} 
        \label{fig:fcase1}
    \end{subfigure}
    \vspace{2pt} 
    \caption{Failure case of CREDiT.}
    \label{fig:all_fcases}
    \vspace{-11pt}
\end{figure}

\subsubsection{Analysis of Disentangled Representations}
The results in Table~\ref{tab:fc_fnc_analysis} provide quantitative evidence for the proposed disentanglement framework. When only the causal representation $F_c$ is used as input, CREDiT maintains strong QA performance across all three benchmarks. In contrast, using only the non-causal representation $F_{nc}$ leads to a substantial performance degradation. These results suggest that answer-relevant information is primarily disentangled in $F_c$, while the information encoded in $F_{nc}$ contributes considerably less to answer prediction.

\subsubsection{Analysis of Successful Cases} 
Fig.~\ref{fig:all_cases} presents a representative successful case. Given the question, CREDiT accurately concentrates on the players and interaction regions related to the action of “steal”, while the baseline attention is more diffuse and partially distracted by irrelevant background regions. This demonstrates that CREDiT can better identify answer-relevant visual evidence for answering, leading to more faithful and interpretable reasoning. Although a few contextual patches are also retained, this is reasonable since moderate scene context is still useful for understanding sports interactions.

\subsubsection{Analysis of Failure Modes} 
In the representative case in Fig.~\ref{fig:all_fcases}, the failure may be attributed to temporal action transitions and domain-specific concepts (e.g., explosive support). Under these conditions, the disentanglement module is restricted by temporal and semantic ambiguity, causing the framework to fall back to diffuse attention similar to the native Qwen backbone. To enhance multi-step reasoning capabilities and broaden practical applicability, integrating domain-specific knowledge and structured reasoning methods presents a promising research direction.

\section{Conclusion}

In this paper, we presented CREDiT, a counterfactual reasoning framework designed to address spurious statistical correlations in VideoQA by disentangling causal visual evidence. By formulating VideoQA from a causality-aware perspective, CREDiT explicitly disentangles cross-modality representations into causal and non-causal components under independence and minimality constraints.
To achieve faithful disentanglement without relying on expensive annotations, we introduce feature-level causal interventions to simulate counterfactual input configurations, enabling the isolation of answer-relevant visual evidence from confounding non-causal context.
Experiments on three benchmarks demonstrate improved QA accuracy and stronger evidence grounding, particularly in sports-domain settings.
For future work, a promising direction is to integrate structured reasoning methods to enhance multi-step reasoning, and to incorporate cross-domain knowledge such as knowledge graphs to further reduce fine-grained semantic ambiguity in real-world scenarios.



\bibliographystyle{IEEEtran}
\bibliography{cite}

\vfill
\end{document}